\documentclass{article}

\usepackage{arxiv}

\usepackage[utf8]{inputenc} 
\usepackage[T1]{fontenc}    
\usepackage{hyperref}       
\usepackage{url}            
\usepackage{booktabs}       
\usepackage{amsfonts}       
\usepackage{nicefrac}       
\usepackage{microtype}      
\usepackage{lipsum}		
\usepackage{graphicx}
\usepackage{natbib}
\usepackage{doi}
\usepackage{amsmath}
\usepackage{fontawesome5}

\title{Schema-Driven Actionable Insight Generation and Smart Recommendation}

\author{ \href{https://orcid.org/0000-0002-3926-6653}{\includegraphics[scale=0.06]{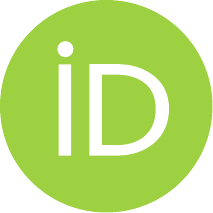}\hspace{1mm}Allmin Susaiyah} \\
	Dept. of Mathematics and Computer Science\\
	Eindhoven University of Technology\\
	Netherlands \\
	\texttt{a.p.s.susaiyah@tue.nl} \\
	 \AND
\href{https://orcid.org/0000-0002-2966-3305}{\includegraphics[scale=0.06]{orcid.pdf}\hspace{1mm}Aki H{\"a}rm{\"a}} \\
	 Advanced Computing Sciences \\
	 Maastricht University\\ The Netherlands \\
	 \texttt{aki.harma@maastrichtuniversity.nl} \\
  \And
\href{https://orcid.org/0000-0002-3261-2920}{\includegraphics[scale=0.06]{orcid.pdf}\hspace{1mm}Milan Petkovi{\'c}} \\
	Dept. of Mathematics and Computer Science\\
	Eindhoven University of Technology\\
	The Netherlands \\
\texttt{m.petkovic@tue.nl} \\
}

\renewcommand{\shorttitle}{\textit{arXiv} Template}

\hypersetup{
pdftitle={Opinion Mining using Population-tuned Generative Language Models},
pdfsubject={q-cs.CL},
pdfauthor={Allmin Susaiyah},
pdfkeywords={Schema, Insight Generation, Insight Recommendation, Neural Networks},
}

\begin{document}
\maketitle

\begin{abstract}
In natural language generation (NLG), insight mining is seen as a data-to-text task, where data is mined for interesting patterns and verbalised into 'insight' statements. An 'over-generate and rank' paradigm is intuitively used to generate such insights. The multidimensionality and subjectivity of this process make it challenging. This paper introduces a schema-driven method to generate actionable insights from data to drive growth and change. It also introduces a technique to rank the insights to align with user interests based on their feedback. We show preliminary qualitative results of the insights generated using our technique and demonstrate its ability to adapt to feedback.
\end{abstract}


\section{Introduction}
An insight is a deep and accurate comprehension of behaviour or patterns observed in data. An actionable insight helps to drive growth and change \cite{jorno2018constitutes}. They can be expressed as a comparison of a measurement made in two distinct yet equivalent contexts, as in \cite{harma2016probabilistic}. For example, one could compare a user's sleep measurements between two comparable contexts say during Mondays and other days. This can be written as, "On Mondays, you sleep less than on the other days". Such insights when communicated to the user help them to understand \cite{Reiterdata2text} or take corrective actions to their behaviour \cite{jorno2018constitutes, o2012data}.

There are three important criteria for an actionable insight: 1) truthfulness \cite{funke2018interactive, agrawal1996parallel, harma2016probabilistic} 2) significance \cite{agrawal1996parallel, harma2016probabilistic} and 3) usefulness \cite{freitas1999rule, fayyad1996data}. The truthfulness of an insight can be asserted using statistical tests and the significance can be asserted using mathematical models. Meanwhile, the usefulness of insight depends on various factors like quantify-ability, meaningfulness, understandability and surprisingness \cite{freitas1999rule,fayyad1996data,ecai-paper,sudarsanam2019rate}. 

The state-of-the-art natural language generation models such as GPTs \cite{gpt2,brown2020language} have limitations when generating insights from large datasets due to their size constraints and hallucinations. Generating actionable insights on a large scale can be done using the generalised framework (GENF) introduced in \cite{susaiyah2020towards} and the 'over-generate and rank' (OGR) paradigm \cite{gatt2018survey}. The GENF framework talks about components that look at (analyse) the data, represent the insights in an intermediary format, say (generate) the statement, obtain user feedback and upgrade the system appropriately. Once many truthful and significant insights are generated using OGR, recommendation algorithms can be used to rank them and select useful insights.


This paper introduces the schema-driven actionable insight-generation approach that grew out of our recent experiences in various applications. The main contribution of this paper is the method to generate actionable insights on a large scale using a controlled natural language input that we call an insight schema. It is a direct implementation of the GENF framework. As a secondary contribution, we present a training protocol that can be used by machine learning models to recommend relevant and diverse insights to the user even with very little feedback. Thirdly, GEN-IG\footnote{\url{https://youtu.be/dAzHRPTloUg}}: A toolkit to perform this along with full documentation is available in {\url{https://github.com/allmin/gen-ig}.  
\begin{figure*}
    \centering
    \includegraphics[width=\textwidth]{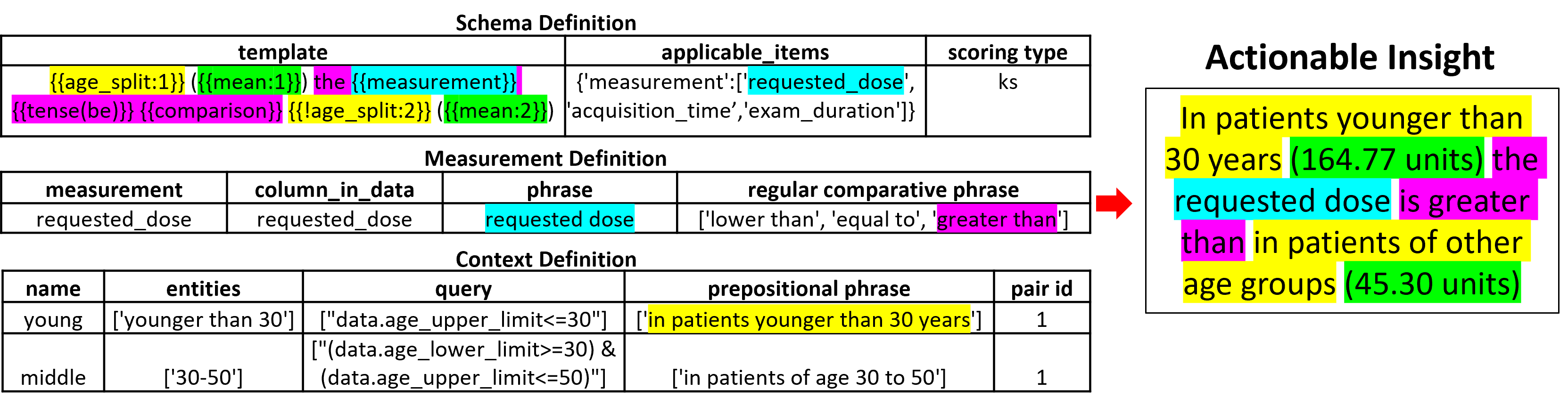}
         \caption{Insight schema definition and a sample insight. The colors indicate corresponding blocks}
         \label{fig:schema}
\end{figure*}
\section{Insight schema}
The insight schemas allow us to define prospective insights using
 a controlled language and generate them complying with the actionability constraints. We developed the GEN-IG toolkit to understand such schemas and generate insights.

\subsection{Insight schema definition}
The 'template' in the schema definition shown in Figure \ref{fig:schema} mentions the data necessary to generate the insights along with the technique to validate them in 'scoring type'. The template compares a continuous variable: 'measurement' across two age groups (categorical variables): 'age\_split:1' and 'age\_split:2'. The eligible 'measurement' for the schema is enumerated in the column 'applicable\_items'. The schema includes mean:1 and mean:2 to display the mean of the 'measurement' in contexts 1 and 2 respectively. This helps in the actionability of such insights. An insight belonging to the above schema is also shown in the figure. The recommended "scoring tests" applicable to different types of schema templates are given below:
\label{sec:truth}
\begin{enumerate}
    \item Two-sample Kolmogorov-Smirnov (KS) test \cite{kstest-naaman2021tight} and the Mann-Whitney U test \cite{mannwhitney-testfay2010wilcoxon}: Used while comparing the distributions of measurement across two contexts.

    \item Binomial test: Used when one of the contexts is a scalar or when the comparison is made on the frequencies of the contexts instead of their distributions.

\end{enumerate}

Other types of statistical tests can be used with our approach depending on what characteristics of the data are being compared.

\subsection{Measurement and context definition}

In addition to the insight schema definitions, the measurement and context definitions help with text realisation and data querying respectively. Figure \ref{fig:schema} shows the measurement definition. The highlighted definition realises "{{measurement}} {{tense(be)}} {{comparison}}" of Figure \ref{fig:schema} into: "the requested dose is greater than" where the verb 'be' takes the right tense and subject-verb agreement based on the contexts and subject-verb agreement. Additionally, a tolerance level $\tau$ is defined which is used to rank the insights (see Section \ref{sec:tolerance}).


 Figure \ref{fig:schema} also shows a sample context definition that specifies, for each contexts, the pandas \cite{reback2020pandas} or SQL queries to extract the data. When two or more contexts are comparable, for example, 'in patients younger than 30 years' and 'in patients of age 30 to 50', they are assigned the same pair id.

\section{Stages of insight generation}
The stages of the proposed approach are elaborated below.

\subsection{Insight library generation}
 It corresponds to the what, where, and how-to-look components of the GENF. Here, individual measurements and contexts are enumerated based on the insight schema definitions. The enumerations are used to create intermediate templates and queries that will be used in the subsequent steps. 

\subsection{Insight scoring}
Scoring the insights helps in recommending them. Each insight is tested for its truthfulness using statistical tests mentioned in Section \ref{sec:truth}. Subsequently, the true insights (p<0.05) are assigned a relevance score based on their completeness, significance defined by the tolerance level \(\tau\)\label{sec:tolerance} and usefulness. The completeness score $Score_C$ as shown in Equation \ref{eq:1} is calculated using the sampling rate of measurements $F_{exp}$, the time-span of the queried data $T$, and the actual recorded data samples $N_{rec}$. The significance score $Score_S$ is calculated using $\gamma$, which determines the margin of error and $\delta$, the calculated difference of means of the measurements across the two contexts. 

A neural network is used to provide a usefulness score $Score_U$ to the insights. For this, user feedbacks: "not useful at all", "not useful", "neutral", "useful", "very useful" are linearly mapped to 0 to 1 as labels to train a modified implementation of the siamese neural network proposed in \cite{susaiyah2021neural}. It uses context means and bag-of-schema words (BoSW) features. The main modification of the network from \cite{susaiyah2021neural} to our network is that our network uses context means instead of histograms. Therefore the input dimensions of our model are much smaller and allow insights involving comparison to scalar values. The BoSW is similar to the bag of words \cite{weinberger2009feature}, but, we pick the context and measurement labels from the schema definition which are more relevant than the surface forms. Neural networks require a lot of training data, hence to adapt to fewer labelled data availability, we follow semi-supervised training \cite{vapnik1974theory}.
Here, we treat the few labelled insights as seed data to turn unlabelled insights into pseudo-labelled insights using a K-nearest neighbours algorithm \cite{knearest} based on the Euclidean distance of the BoSW features. The pseudo-labelled data is then used to train the network model. In the absence of user feedback, the $Score_U$ is assigned a value of 1. The overall relevance score of insight is calculated as a product of the completeness, significance and usefulness scores.
    \begin{equation}
        \label{eq:1}
        \begin{gathered}
        Score_C = \frac{N_{rec}}{F_{exp} * T} \\
        Score_S = \frac{1}{1 + \exp(-\frac{\gamma\delta}{\tau})} \\
        Score_F = Score_C * Score_S * Score_U
        \end{gathered}
    \end{equation}
    
\subsection{Surface realisation}
This stage corresponds to the how-to-say component of the GENF. This leads to a complete insight text as shown in Figure \ref{fig:schema}. Here, placeholders like $\{context:1\}$, $\{mean:1\}$, etc are filled in. The tenses of verbs are modified based on the context. For example, if the context of a verb relates to a period in the past, the past tense is assigned to it. The verbs that have to be modified are mentioned in the template of the form $\{tense(verb,x)\}$, where x is 2 if the verb indicates the user (second person) and 3 (third person) if it indicates the measurement.

\subsection{Insight recommendation}
The insights populated are usually in large numbers, and hence, a selection mechanism is employed in our approach to choosing the best and most diverse set of insights (what-to-say). Here, the insights are clustered based on BoSW features. For clustering, the K-means algorithm \cite{kmeans} based on Euclidean distance is used. The value of K is set to be equal to the number of insights that we want to present to the user. Finally, we choose to show to the user, the insight having the highest \(Score_F\) from each cluster.

\section{Experiments}
\subsection{Preliminary insights and feedback on usefulness}
We used our approach and generated insights on the utilization data of an interventional radiology department of a hospital [Anonymous]. The schemas consider contexts like physicians, exams, periods and patient age over different measurements like exam duration, acquisition time for a scan, radiation dosage and physician shift times. 
A total of 2470 insight candidates were generated of which 730 were truthful and scored using a $\gamma$ value of 6. $Score_U$ is assigned 1 due to the absence of feedback. A 51-element BoSW feature vector is used to select 23 insights by the clustering approach. We then collected feedback from the hospital administration on the usefulness of the insights. A few of these insights and the feedback obtained are shown in Table \ref{tab:before_feedback}.

\begin{table}
    \centering
    \caption{Sample set of insights recommended and feedback obtained (FB):\faThumbsUp: useful, \faThumbsDown: not-useful}
    \begin{tabular}{p{6.3cm}|p{0.5cm}}\hline
    
Insight & FB \\ \hline
With physician-9 (0.25 hours) the duration of a shift was lower than other physicians (4.27 hours)
 &  \faThumbsUp\\ \hline
 On Wednesdays, the duration of an exam in the morning (1.37 hours) is greater than in the afternoon (0.74 hours)
 & \faThumbsUp\\ \hline
 On Thursday mornings (10.03) the acquisition time of an exam was 3.92\% lower than on other day mornings (10.44) & \faThumbsDown\\ \hline

    \end{tabular}
    
    \label{tab:before_feedback}
\end{table}
\begin{figure}
    \centering
    \includegraphics[width=0.45\textwidth]{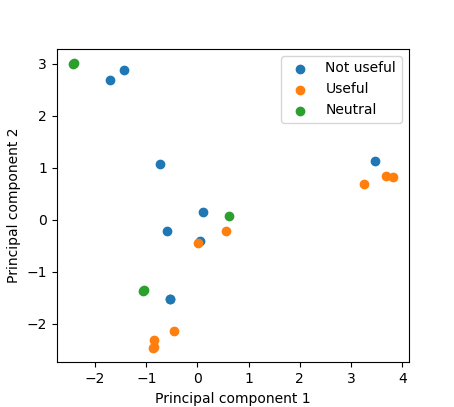}
    \caption{Principal component projections of the insights}
    \label{fig:exp1}
\end{figure}
\begin{table}
    \centering
    \caption{Insights recommended by our selection approach after one round of feedback from the user}
    \begin{tabular}{p{6cm}|p{0.5cm}}\hline
    
Insight & $Score_U$\\ \hline
With physician-20 (6.55 exams) the number of exams in a shift was greater than other physicians (3.61 exams) & 0.67\\ \hline
 On Wednesday afternoons the duration of an exam in patients of age 50 to 70 (29.51 minutes) is lower than in patients older than 70 years (51.75 minutes)
 & 0.67\\ \hline
 On Wednesday afternoons (9.56 s) the acquisition time of an exam was 13.55\% lower than on other day afternoons (11.05 s)
 & 0.8\\ \hline
    \end{tabular}
    \label{tab:after_feedback}
\end{table}
\begin{table}
    \centering
    \caption{Statistics of the feedback given to the insights \\
    $\diamondsuit$: First round of feedback (default recommender) \\
    $\heartsuit$: Second round of feedback (neural recommender)}
    \begin{tabular}{p{3cm}|p{1cm}|p{1cm}}\hline
        Statistic & $\diamondsuit$& $\heartsuit$\\ \hline
         Total Insights  & 23 & 23  \\
         not at all useful & 0 & 0 \\
         not useful & \textbf{9} & 8 \\
         neutral & 5 & 6 \\
         useful & \textbf{9} & \textbf{9} \\
         very useful & 0 & 0 \\
         \hline
    \end{tabular}
    \label{tab:stats}
\end{table}
\begin{table}

    \centering
        \caption{Performance of parallel vs serial processing}
    \begin{tabular}{p{1.5cm}|p{1.8cm}|p{1.4cm}|p{1.4cm}}\hline
        Total insights (count) & Significant insights (count) & Time for serial (min) & Time for parallel (min)\\\hline
         615  & 176 & \textbf{2.53} & 2.76 \\
         1237 & 345 & 3.59 & \textbf{3.12} \\ 
         1855 & 554 & 4.65 & \textbf{3.68} \\
         2470 & 730 & 5.86 & \textbf{4.67} \\\hline
    \end{tabular}

    \label{tab:eff}
\end{table}

The two most dominant eigenvectors \cite{wold1987principal} of the insight's BoSW features are shown in Figure \ref{fig:exp1}. It is observed that all the 'not useful' insights seem to be clustered together. This indicates the discriminating power of the BoSW features. The statistics of the feedback are shown in Table \ref{tab:stats}. It is seen that quite a few insights are not-useful. This can be further improved by incorporating user feedback while recommending the insights. The insights chosen after one round of feedback from the user incorporating the neural network are shown in Table \ref{tab:after_feedback}. The corresponding statistics are shown in Table \ref{tab:stats}. Although there aren't any significant differences in the usefulness. We observed a slight drop in the number of insights that were not useful which could turn significant with multiple rounds of feedback.

\section{Discussions}
\subsection{Scalability}
Scalability is important for the effective implementation of the OGR paradigm. Schematising the insight generation divides the problem into independent subsets that allow for parallel processing. A comparison of the computation time with and without parallel processing is presented in Table \ref{tab:eff}. The rate of generating insights by the parallel implementation of GEN-IG is slower than its serial counterpart when the number of insights is less than 700. However, this changes when the insights increase in numbers.


\subsection{Domain-adaptation}
The schema approach introduced here can be extended to other domains by adapting the schema instead of re-coding. Additionally, the generalisation capabilities of large language models could be leveraged to perform this adaptation task in future.

\section{Conclusion}
In this paper, we presented the schema-based approach that allows us to easily define, control and generate actionable insights. We also described a recommendation system that picks diverse, significant and useful insights to be shown to the user using features derived from the schema. In our preliminary experiment, we found a slight improvement in the usefulness of the insights with one round of feedback which we believe would improve further with subsequent rounds. In future, we plan to implement this approach in more frequent feedback scenarios such as personal health coaching and campaign management. 

 \section*{Acknowledgements}
 This work is supported by the Horizon H2020 Marie Skłodowska-Curie Actions Initial Training Network European Industrial Doctorates project under grant agreement No. 812882 (PhilHumans).


\bibliographystyle{unsrtnat}
\bibliography{insightgeneration.bib}  






\end{document}